\documentclass{article}

\usepackage{graphicx}
\usepackage{booktabs}
\usepackage{makecell}
\usepackage{amsmath}
\usepackage[authoryear,sort&compress,round]{natbib}
\usepackage{hyperref}
\usepackage{diagbox}
\usepackage[usenames,dvipsnames]{color}
\hypersetup{
  colorlinks,
  citecolor=Blue,
  linkcolor=Red,
  urlcolor=Green}

\usepackage[final]{neurips_2021}

\usepackage[utf8]{inputenc} 
\usepackage[T1]{fontenc}    
\usepackage{hyperref}       
\usepackage{url}            
\usepackage{booktabs}       
\usepackage{amsfonts}       
\usepackage{nicefrac}       
\usepackage{microtype}      
\usepackage{xcolor}         

\title{Oriented Object Detection with Transformer}

\author{Teli Ma\textsuperscript{1,${\dag}$}, 
 Mingyuan Mao\textsuperscript{2,${\dag}$}, 
 Honghui Zheng\textsuperscript{3}, 
 Peng Gao\textsuperscript{3}, 
 Xiaodi Wang\textsuperscript{3}, 
 Shumin Han\textsuperscript{3,*}, \\
 \textbf{
 Errui Ding\textsuperscript{3},
 Baochang Zhang\textsuperscript{2},
 David Doermann\textsuperscript{1,*}}\\

\textsuperscript{1}University at Buffalo, Buffalo, USA \\
\textsuperscript{2}Beihang University, Beijing, China\\
\textsuperscript{3}Department of Computer Vision Technology (VIS), Baidu Inc\\

\textsuperscript{*}Corresponding author, email: hanshumin@baidu.com, doermann@buffalo.edu\\
\textsuperscript{${\dag}$}Equal contributions
}

\begin{document}

\maketitle

\begin{abstract}
Object detection with Transformers (DETR)  has achieved  a competitive performance over traditional detectors, such as Faster R-CNN. However, the potential  of DETR remains largely unexplored  for the more challenging task of arbitrary-oriented object detection problem. We provide the first attempt and implement Oriented Object DEtection with TRansformer ($\bf O^2DETR$) based on an end-to-end network. The contributions of $\rm O^2DETR$ include: 
1) we provide a new insight into oriented object detection, by applying Transformer to directly and efficiently localize objects without a tedious process of rotated anchors as in conventional detectors; 2) we design a simple but highly efficient encoder for Transformer  by replacing the attention mechanism with depthwise separable convolution, which can significantly reduce  the memory and computational cost of using multi-scale features in the original Transformer; 3) our $\rm O^2DETR$ can  be another new benchmark in the field of  oriented object detection, which achieves up to 3.85 mAP improvement over Faster R-CNN and RetinaNet. We simply fine-tune the head mounted on $\rm O^2DETR$ in a cascaded architecture and achieve a competitive performance over SOTA in the DOTA dataset. 

\end{abstract}

\section{Introduction}

Arbitrary-oriented targets are widely distributed in application scenarios like scene text detection and remote sensing object detection~\citep{xia2018dota}. 
Detecting oriented targets with anchors without rotation is difficult as targets are always tiny, oblique and densely packed. 
Based on that, many rotated detectors like $\rm R^3Det$~\citep{yang2019r3det}, $\rm S^2ANet$~\citep{han2020align}, ReDet~\citep{han2021redet} are proposed to detect objects based on traditional detectors like Faster R-CNN~\citep{ren2016faster}, RetinaNet~\citep{lin2017focal} by adding rotation of pre-set anchors while learning. However, the rotated-anchor regression and post-process like non-maximum suppression are based on a tedious and redundant process,  which is an indirect and sub-optimal solution to the oriented object detection problem.

\begin{figure}
    \centering
    \includegraphics[width=1\linewidth]{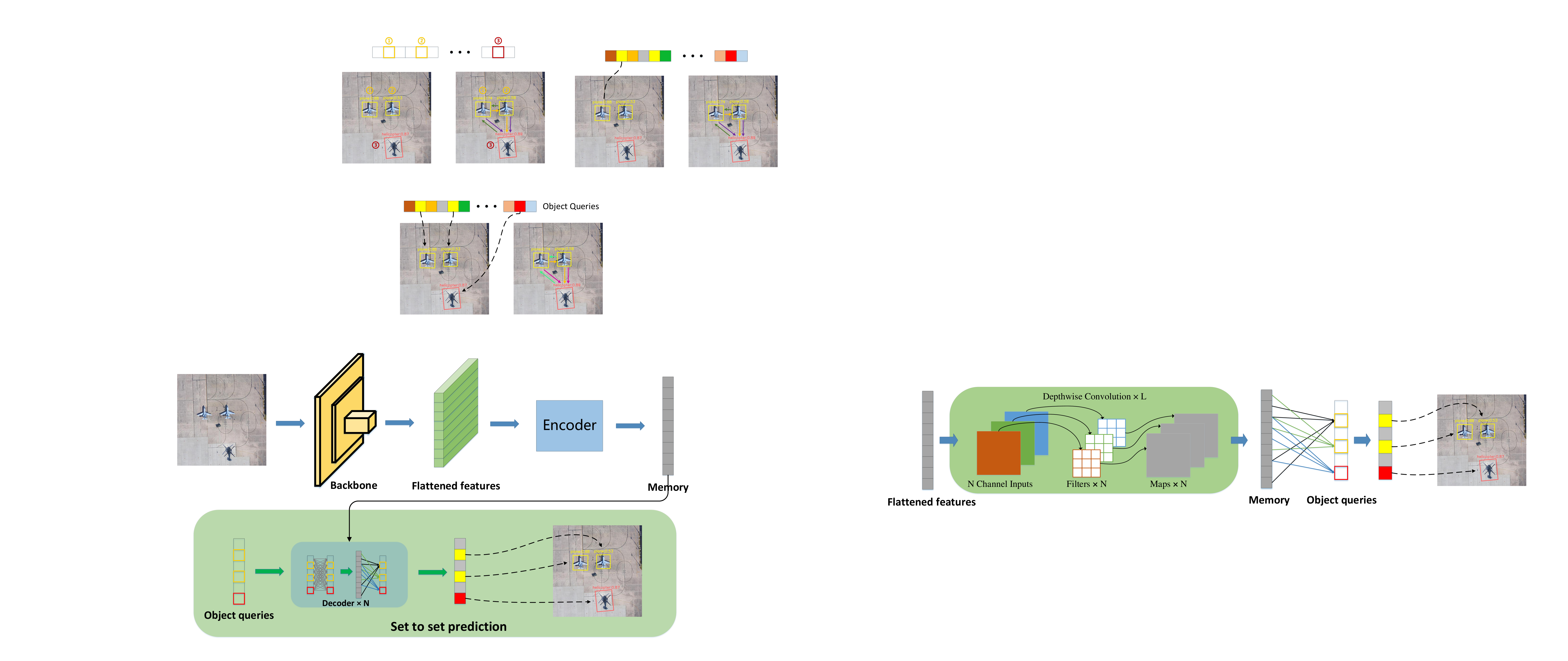}
    \caption{The end-to-end scheme of  $\rm O^2DETR$}
    \label{fig1}
\end{figure}

In this paper, based on Transformer~\citep{vaswani2017attention} and DETR~\citep{carion2020end}, we introduce an  Oriented Object DEtection with TRansformer ($\bf O^2DETR$) method, which is the first attempt to apply  Transformer to the oriented object detection task.
We provide a direct method for  oriented object  detection by matching angled boxes with oriented objects end-to-end as shown in Fig.~\ref{fig1}. Specifically, we pre-set and improve the fixed-length object queries with angle dimension to interact with the encoded features, and extract angle-dimensional information via cross-attention mechanism. The set of angle-aware object queries match ground truths with bipartite matching during training.

Considering significant scale variances for different categories of oriented objects (large ones like ground-track-field and tiny ones like small-vehicle), multi-scale feature maps are necessary for object detection. However, global reasoning scheme of attention mechanism of the original Transformer encoder is highly and computationally complex for multi-scale features. Furthermore, we argue that  global reasoning is actually not necessary, especially when oriented objects of the same category are always densely packed and the object query only interacts with visual features around the object rather than those of whole global image. Based on these observation, we are inspired to introduce local aggregation with depthwise separable convolutions which can perform 
much better than the original self-attention mechanism of Tranformer. 
Replacing with convolutions shortens the Transformer training epochs and achieves a fast convergence compared with the conventional attention mechanism, because  the information  exchange only happens  among adjacent pixels when extracting features.

\begin{figure}
    \centering
    \includegraphics[width=1\linewidth]{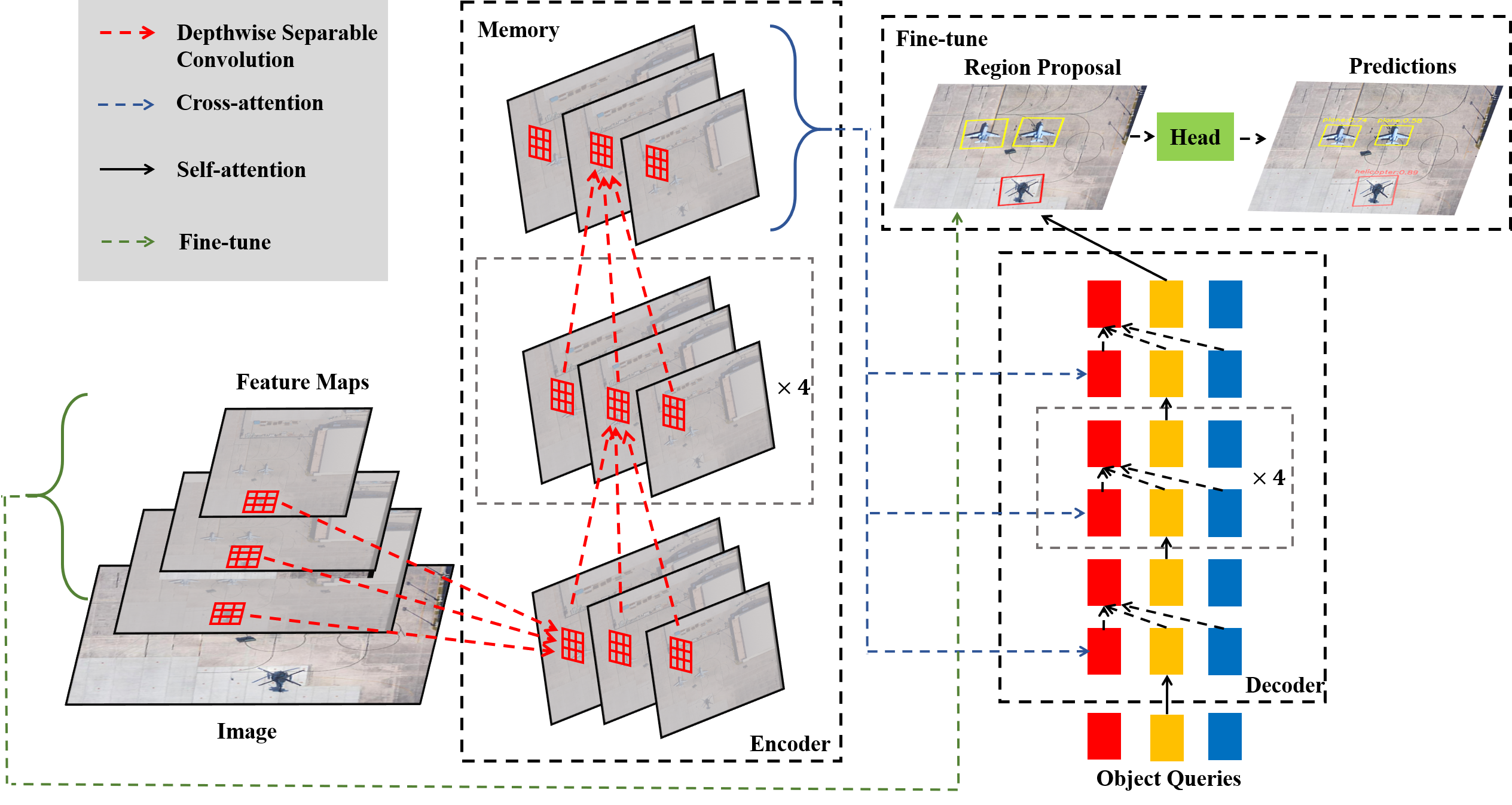}
    \caption{Illustration of the proposed $\rm O^2DETR$ and fine-tune method.}
    \label{fig2}
\end{figure}

Experiments on DOTA~\citep{xia2018dota} dataset demonstrate our $\rm O^2DETR$ outperforms both one-stage and two-stage rotated detectors without any refinement by $3.85\%$ mAP. Based on that, we fine-tune the head mounted on  $\rm O^2DETR$ with a cascaded refinement module to boost performance of our detector further. With parameters of Transformer fixed, we use predictions of $\rm O^2DETR$ as the region proposal and  
select features from feature maps generated by backbone to fine-tune a cascaded prediction head as shown in Fig.~\ref{fig2}. We only fine-tune the head  and achieve $79.66\%$ mAP on DOTA test dataset with backbone ResnNet-50, leading to a competitive performance compared with the state-of-the-art.

\section{Related Works}

\textbf{Oriented Object Detection.}
Oriented object detection refers to building detectors using rotated bounding box representation. Arbitrary-oriented targets are widely distributed in remote sensing and text images. These targets are often crowded, distribute with large scale variations and appear at arbitrary orientations~\citep{xia2018dota}. Thus, existing methods built on detectors using horizontal bounding boxes suffer from containing several objects of interest in one anchor/RoI. Some methods are adopted to alleviate the problem. R-RPN~\citep{ma2018arbitrary} uses rotated region proposal networks to detect oriented targets by rotated proposals. In R$^{2}$CNN~\citep{jiang2017r2cnn}, a horizontal region of interest (RoI) is leveraged to predict both horizontal and rotated boxes. RoI Transformer~\citep{ding2019learning} transforms the horizontal RoI in R$^{2}$CNN into a rotated one (RRoI). SCRDet~\citep{yang2019scrdet} and RSDet~\citep{zhou2020objects} focus on boundary problem caused by periodicity of angle and propose novel losses to fix it. In CSL~\citep{yang2020arbitrary}, angle regression is converted into classifying accurate angle value in one period. R$^{3}$Det~\citep{yang2019r3det} samples features from center and corners of the corresponding anchor box and sum them up to re-encode the position information in order to solve misalignment of classification and localization. S$^{2}$A-Net~\citep{han2020align} uses aligned convolution network to refine the rotated boxes and align features. Recently proposed ReDet~\citep{han2021redet} incorporates rotation-equivariant network into detector to extract rotation-equivariant features. To the best of our knowledge, all previous rotated detectors make predictions in a indirect way of rotating anchors or proposals rather than predicting with angle knowledge directly as our $\rm O^2DETR$.

\textbf{Transformers.}
Transformers~\citep{vaswani2017attention} include both self-attention and cross-attention mechanism and achieve success in not only machine translation~\cite{ott2018scaling,gao2020multi}, but also model pretraining~\citep{devlin2018bert, radford2018improving, radford2019language, brown2020language}, visual recognition~\citep{ramachandran2019stand, dosovitskiy2020image,mao2021dual} and multi-modality fusion~\citep{yu2019deep, lu2019vilbert, gao2019dynamic, gao2019multi, geng2020dynamic}. Transformers perform information exchange between all sets of inputs using key-query value attention. The complexity of information exchange hinders model scalability in many cases for limiting input sequences. Many methods have been proposed to solve the problem. Reformer~\citep{kitaev2020reformer} proposes a reversible FFN and
clustering self-attention. Linformer~\citep{wang2020linformer} and FastTransformer~\citep{katharopoulos2020transformers} propose to remove the softmax in the transformer and perform matrix multiplication between query and value first to obtain a linear-complexity transformer. Adaptive Clustering Transformer(ACT)~\citep{zheng2020end} perform an approximated self-attention by clustering key and query feature. 
LongFormer~\citep{beltagy2020longformer} perform self-attention within a local window instead of the whole input sequence. In Deformable DETR~\citep{zhu2020deformable}, attention mechanism works on limited sample points rather than all image pixels, thus reduces training epochs largely compared with DETR~\citep{carion2020end}. In SMCA-DETR~\citep{gao2021fast}, a spatially-modulated Gaussian mechanism has been introduced to coupled attention map with the position of predicted bounding-box and achieve fast convergence speed compared with DETR. In our model, we utilize depthwise separable convolutions to replace self-attention mechanism used in encoder to speed up training and save memory.

\textbf{Depthwise Separable Convolution.}
Depthwise separable convolutions were first studied by~\citep{sifre2013rotation} from Google Brain. In 2016, it was demonstrated a great success on large-scale image classification in Xception~\citep{chollet2017xception}. Later, the depthwise separable convolutions proved to reduce the number of parameters of models (the MobileNets family of architectures~\citep{howard2017mobilenets}) considerably.
A depthwise separable convolution consists in a depthwise convolution, i.e. a spatial convolution performed independently over each channel of an input, follewed by a pointwise convolution, i.e. a $1\times1$ convolution projecting the channels output of the depthwise convolution onto a new channel space. Previous work on depth-wise research focus on the light-weight characterstic. Our research show that the strong contextual aggregation ability of depthwise convolution than strong models like attention and deformable attention.

\section{Method}
\label{secmethod}
$\rm O^2DETR$ is the first Transformer baseline in the  oriented object detection domain to our best  knowledge. Different from detecting oriented objects by rotating pre-set anchors to match objects, $\rm O^2DETR$ predicts rotated boxes directly from a set of object queries. 
Our work can be concluded as two points:
\begin{itemize}
    \item Utilize separable depthwise convolutions to replace attention mechanism for an efficient  Transformer encoder.
    \item Take advantage of high recall rate achieved by our $\rm O^2DETR$ to fine-tune the baseline for a better performance.
\end{itemize}
In the rest of the section, we discuss the setup process in details to show the process of constructing the model step by step.

\subsection{Depthwise Separable Convolution and Attention}
 The original attention mechanism is computationally complex when processing multi-scale feature maps due to the global reasoning scheme, which is designed to capture relationships among all feature points in a global image. Differently, the depthwise convolution is  proposed to for light-weight characteristic research, which shows the strong contextual aggregation and visual expression ability. The depthwise separable convolution operation consists of a depthwise convolution, which is a spatial convolution performed independently over every channel of the input, followed by a pointwise convolution with $1\times1$ filters projecting the channels computed by former depthwise convolution into new channel space. The  formulation of a convolution operation is given as: 
\begin{equation}
    Conv(W, y)_{(i,j)} = \sum_{k,l,c}^{K,L,C}W_{(k,l,c)}\cdot y_{(i+k, j+l, c)},
\end{equation}
 where $K,L$ denote kernel size of convolution filter and the $y_{(i+k, j+l, c)}\in \mathbb{R}^{1}$ denotes a spatial feature point on c$th$ ($c\in C$, input feature dimensions) channel. Then the depthwise separable convolution ($DSConv$ for abbreviation)  can be formulated as:
\begin{equation}
\label{eq2}
\begin{split}
     Depthwise&Conv(W,y)_{(i,j)} = \sum_{k,l}^{K,L}W_{(k,l)}\odot y_{(i+k, j+l)}, \\
     Pointwise&Conv(W,y)_{(i,j)} = \sum_{c}^{C}W_{c} \cdot y_{(i,j,c)}, \\
     DSConv (W_p, W_d, y)_{(i,j)} &= PointwiseConv_{(i,j)}(W_p, DepthwiseConv_{(i,j)}(W_d, y)),
\end{split}
\end{equation}
 where $\odot$ means the element-wise product and $y_{(i+k, j+l)} \in \mathbb{R}^C$ denotes a feature point in all channels. $DepthwiseConv(W,y)_{(i,j)}\in \mathbb{R}^{(H,W)\times C}$ denotes feature space generated by depthwise convolution and $PointwiseConv(W,y)_{(i,j)}$ denotes spatial feature space after utilizing pointwise convolution on results of depthwise convolution. 

 From Eq.~\ref{eq2} we can see the core idea of depthwise separable convolution which divides the feature learning into two separate steps, one is the spatial feature learning and the other is channel interaction. Then essence of such mechanism is to flatten and weight features of each channel. By comparison,  the attention mechanism of Transformer consists of information exchange between the set of query and key elements. Given a set of query elements and a set of key elements, the attention mechanism adaptively aggregate the key contents according to the attention weights based on the measurement of compatibility of query-key pairs. The visual features are captured according to the attention weights, resulting in visual feature points interacting with others in the global image space. Let $\Omega_q\in \mathbb{R}^{H\times W}$ and $\Omega_k\in \mathbb{R}^{H\times W}$ represent the set of query and key elements, when $q\in \Omega_q$, $k\in \Omega_k$ index a query element wih representation feature $y_q\in \mathbb{R}^C$ and a key element with representation feature $y_k\in \mathbb{R}^C$. The attention feature is calculated as follow:
 \begin{equation}
     Attn(y_q,\textbf{y}_{\Omega_k})=\sum_{k\in \Omega_k} A_{(q,k)} \cdot W \odot y_k,
 \end{equation}
 where $W\in \mathbb{R}^{C}$ is of learnable weights and attention weights $A_{qk}$ are normalized as $\sum_{k\in \Omega_k}A_{(q,k)}=1$. In the visual Transformer filed, the representation features $y_q$ and $y_k$ are usually of the concatenation/summation of element contents and positional embedding for $2D$ positional meaning. $y_q$ and $y_k$ are visual feature points in all channels with position embedding and apply dot multiplication, which is more similar to depthwise convolution rather than conventional convolution.
 
 The depthwsie separable convolution is more efficient than  attention mainly in the sampling space of visual feature points. To every query feature $y_q$, $Attn(y_q,x)$ do visual information interaction with all key features through attention weights $A_{(q,k)}$, while the  $W_{k,l}$ of depthwise convolution works on every feature map point $y_{(i,j)}$ to interact with local feature points around. We hypothesize the local aggregation of depthwise  convolution performs better on tiny and dense objects, avoiding long training schedules before convergence of global feature interaction. Meanwhile, the depthwise separable convolutions reduce the complexity of model compared with attention mechanism. Suppose the channels, width, height of a feature map as $C, W, H$, the complexity of attention mechanism is $\mathcal{O}(HWC^2)$, while the complexity of depthwise separable convolution turns out to be $\mathcal{O}(kC+C^2)$ ($k$ is the kernel size of filter and $k<<HW$). It is safe to say replacing attention mechanism with depthwise separable convolution could save parameters and fasten training.

 \subsection{$\rm \bf{O^2DETR}$}
 \paragraph{Multi-scale Feature Presentation.}
 Most oriented object detectors utilize multi-scale feature maps due to large variance of objects scales. Our proposed $\rm O^2DETR$ adopts multi-scale feature maps generated by backbone to enrich visual feature  presentation. Given an input image, the encoder extracts the multi-scale visual features from the output feature maps $\{x^l\}_{l=1}^{L-1}$  of stages $C_3$ through $C_5$ in ResNet~\citep{he2016deep}, where $C_l$ is of resolution $2^l$ lower than the input image. The lowest resolution feature map $x^L$ is obtained via a $3\times 3$ and stride $2$ convolution on the final $C_5$ stage, and denoted as $C_6$. All the feature maps of different scales are of 256 channels.
Transformer encoder encodes all locations of different scales in multi-scale feature maps by propagating and aggregating information between pixels of different scales. The large number of pixels demonstrates the feasibility of replacing attention mechanism with convolution in encoding tokens. Specifically, we sum features from the adjacent levels into each level to fuse the different scale features for information fluid to acquire more semantic information, which is formulated as
 \begin{equation}
 \begin{split}
     DSConv(W_p, W_d, x)^l = DSConv(W_p, W_d, x)^l + &\textbf{Dropout}(DSConv(W_p, W_d, x)^{l-1} \\ + DSConv(W_p, W_d, x)^{l+1}), \quad l\in [1, L-1].
 \end{split}
 \end{equation}
 Given the encoded multi-scale features $E_{l}$ ($l\in L$), multi-scale cross-attention is conducted between object query and feature maps. For each object query, a $2D$ normalized coordinate of the reference point $p$ is predicted from original object query embedding calculated by linear project layers. The object queries extract multi-scale features from the encoder memory as 
 \begin{equation}
     MSAttn(z_q,p,\{E_l\}_{l=1}^L)=\sum_{l=1}^L\sum_{k\in \Omega_k} A_{(q,k,l)} \cdot W \cdot x_k^l,
 \end{equation}
 where visual features are taken from multiple levels of feature maps and $x_k^l$ is feature point from $E_l$. $p$ is the reference point where decoder extracts image features from. We modify the representation of $p$ by adding an extra angle dimension to estimate original position and angle of original object query as $p_{(c, w, h, \alpha)}$ ($c$ is the center point, $w, h, \alpha$ are width, height and angle of estimation of object queries). The object queries then will be fed into detection head for detection task.

 \paragraph{Detection Head.}
 After conducting cross-attention between the object query and the encoded image features, we can obtain the updated features $D\in \mathbb{R}^{N\times C}$ ($N$ is the length of object queries). In the detection head, a 3-layer MLP and a linear layer are used to predict the bounding box and classification confidence. Different from original detection head in DETR~\citep{carion2020end}, as for bounding boxes, we project the features $D$ into 5-dimensional boxes including center point $x_c.y_c$, width and height $w, h$ and the angle of bounding box $\alpha$. We denote the prediction as
 \begin{equation}
 \begin{split}
     Box_{\{x_c,y_c,w,h,\alpha\}}&=\textbf{Sigmoid}(\textbf{MLP}(D)), \\
     Score&=\textbf{FC}(D).
\end{split}
 \end{equation}
 
 \subsection{Fine-tune $\rm \bf{O^2DETR}$}
 The $\rm {O^2DETR}$ can be  a new baseline model replacing Faster R-CNN~\citep{ren2016faster} and RetinaNet~\citep{lin2017focal} for the oriented object detection problem. As is known, many methods are introduced into the  two baseline models to refine detectors and refresh the SOTA performance. The $\rm {O^2DETR}$ opens up great possibilities of exploiting advantages of end-to-end oriented object detectors, thanks to its simple architecture-construction process, multi-scale visual expression and fast convergence. Motivated by this, we  provide an insight of improving performance of our new Transformer baseline by a simple yet effective  fine-tuning strategy. 
 
 Inspired by the observation of high recall rate of $\rm {O^2DETR}$ revealed in Table~\ref{tab3}, we establish a fine-tune network by exploiting $\rm {O^2DETR}$ as a region proposal generator. To save memory and computational cost, we freeze the parameters of $\rm {O^2DETR}$, just fine-tune an additional prediction head for final bounding boxes and confidence scores. We regard the inferred bounding boxes of $\rm {O^2DETR}$ as region proposals,  utilizing an Region of Interest Align (ROIAlign) network to project the proposals into feature maps obtained from backbone.  The features aligned by ROIAlign would be fed into a prediction head for more accurate predictions of boxes and scores. The process can be formulated as
 \begin{equation}
 \begin{split}
    F = \textbf{ROI}&\textbf{Align}(P, \{x^l\}_{l=1}^{L-1}),\\
    Box_{\{x_c,y_c,w,h,\alpha\}}^F&=\textbf{Sigmoid}(\textbf{MLP}(F)), \\
    Score^F &= \textbf{FC}(F), \\
    \textbf{Box} =  Box&_{\{x_c,y_c,w,h,\alpha\}} + Box_{\{x_c,y_c,w,h,\alpha\}}^F,
\end{split}
 \end{equation}
 where the $F$ is the fine-tuned features and $Box_{\{x_c,y_c,w,h,\alpha\}}^F$, $Score^F$ are predictions of fine-tuned features. We calculate the bounding boxes of fine-tuned features as the residual of original location estimation and add it into the original bounding boxes generated by $\rm O^2DETR$.
 No NMS is applied before feeding the region proposals to the ROIAlign. The fine-tune process is illustrated in Fig.~\ref{fig2}.

\section{Experiments}
\label{secexp}
\paragraph{Dataset.}
We conduct experiments on DOTA~\citep{xia2018dota}, which is the benchmark dataset of oriented object detection. DOTA contains 2806 aerial images with the size ranges from $800\times 800$ to $4000\times 4000$ and 188282 instances with different scales, orientations and shapes of 15 common object categories, which includes: Plane (PL), Baseball diamond (BD), Bridge (BR), Ground track field (GTF), Small vehicle (SV), Large vehicle (LV), Ship (SH), Tennis court (TC), Basketball court (BC), Storage tank (ST), Soccer-ball field (SBF), Roundabout (RA), Harbor (HA), Swimming pool (SP), and Helicopter (HC). The fully annotated DOTA are divided into three parts: half of the images are randomly selected as training set, $1/6$ as the validation set and $1/3$ as the testing set. We crop original images into $1024\times 1024$ patches with a stride of 824. We only adopt random horizontal flipping during training to avoid over-fitting and no other tricks are utilized if not specified.

\paragraph{Implentation Details.}
ImageNet~\citep{deng2009imagenet} pre-trained ResNet-50~\citep{he2016deep} is utilized as the backbone for ablations. In ablations, we denote models extracting features with ResNet-50 and ResNet-101 as $\rm O^2DETR$-R50 and $\rm O^2DETR$-R101, respectively. The fine-tuned model is denoted as F-$\rm O^2DETR$. Multi-scale feature maps are extracted without FPN~\citep{lin2017feature}. We use downsampling ratio of $64,32,16,8$ to process feature maps by default. We set the length of object queries as $1000$ as objects are always dense in DOTA image. 

Performance trained for 50 epochs are reported and the learning rate drops to $1/10$ of its original value at the $40th$ epoch. When training the $\rm O^2DETR$, the learning rate is set as $10^{-4}$ for the Transformer encoder-decoder and $10^{-5}$ for the pre-trained ResNet backbone. $\rm O^2DETR$ is trained by minimizing the classification loss, bounding box L1 loss, and IoU loss with coefficients $2,5,2$, respectively. In Transformer layers, post-normalization is adopted. We use random crop in training with the largest width or height set as 1024 for all ablations. During fine-tuning, we train the model for 1x (12 epochs) by default.
All models are trained on NVIDIA Tesla $8\times$V100 GPUs with 2 images per GPU.

\subsection{Comparison with Faster R-CNN and RetinaNet}
Rotated Faster R-CNN~\citep{ren2016faster} and RetinaNet~\citep{lin2017focal} are the most popular two-stage and one-stage baseline used in oriented object detection domain. Despite the differences in training scheme of anchor-based detectors and Transformer ones, we attempt to compare our $\rm O^2DETR$ baseline with them in a relative fair way. To align with $\rm O^2DETR$, we train the rotated Faster R-CNN and RetinaNet for 3x schedule. Results are report in Table~\ref{tab1}. Data augmentation is not used in all baselines. To be comparable, our $\rm O^2DETR$ uses 6 encoder layers and 6 decoder layers with around 41M parameters using multi-scale features and ResNet-50. To the best of our knowledge, no rotated Transformer baseline has been presented in oriented object detection problem. $\rm O^2DETR$ has better performance in mAP compared with anchor-based baseline detectors when the number of parameters is similar. In conclusion, our proposed $\rm O^2DETR$ could serve as a strong and competitive baseline for oriented object detection problem.

\begin{table}[h]
\begin{center}
\begin{tabular}{|l|lllllllll}
\toprule
Method       & backbone & \multicolumn{5}{l}{MS.dowsnsample ratios}                                                                & Epochs & params  & mAP \\ \cline{3-7}
             &          & \multicolumn{1}{l|}{64} & \multicolumn{1}{l|}{32} & \multicolumn{1}{l|}{16} & \multicolumn{1}{l|}{8} & 4 &                &       &     \\ \midrule
Faster R-CNN & ResNet-50         & &           &       &         &   & \makecell[c]{50}       &\makecell[c]{39M}        & 60.32    \\
~\citep{ren2016faster}             & ResNet-50         & \checkmark                       & \checkmark                      &  \checkmark                       &  \checkmark                       &    &  \makecell[c]{50}    & \makecell[c]{42M}             & 64.17    \\
             &  ResNet-50         &  \checkmark                       &   \checkmark                      & \checkmark                        &  \checkmark                      &\checkmark   & \makecell[c]{50}     &  \makecell[c]{43M}            & 66.25    \\
             &   ResNet-101        &                &           &              &                        &    &\makecell[c]{50}        &  \makecell[c]{60M}       & 62.44    \\       
             &   ResNet-101        &   \checkmark                      & \checkmark                        & \checkmark                        &  \checkmark                      &    &\makecell[c]{50}        &  \makecell[c]{63M}          &  66.03   \\
             &  ResNet-101         &  \checkmark                       & \checkmark                        &  \checkmark                       &  \checkmark                      & \checkmark  & \makecell[c]{50}      &  \makecell[c]{64M}             & 67.71    \\
              \midrule
RetinaNet   & ResNet-50         &                  &             &            &                        &   & \makecell[c]{50}        & \makecell[c]{34M}   & 58.54    \\
~\citep{lin2017focal}             & ResNet-50         & \checkmark                       & \checkmark                      &  \checkmark                       &  \checkmark                       &    & \makecell[c]{50}      &   \makecell[c]{37M}         & 62.78    \\
             &  ResNet-50         &  \checkmark                       &   \checkmark                      & \checkmark                        &  \checkmark                      &\checkmark   & \makecell[c]{50}       &  \makecell[c]{38M}        & 65.77    \\
             &   ResNet-101        &               &          &                        &                     &    & \makecell[c]{50}        &  \makecell[c]{55M}     & 60.47    \\            
             &   ResNet-101        &   \checkmark                      & \checkmark                        & \checkmark                        &  \checkmark                      &    &\makecell[c]{50}        & \makecell[c]{58M}         &  64.11   \\
             &  ResNet-101         &  \checkmark                       & \checkmark                        &  \checkmark                       &  \checkmark                      & \checkmark  &  \makecell[c]{50}       &  \makecell[c]{59M}       &  66.53   \\ \midrule
$\rm O^2DETR$       & ResNet-50         &            &             &                         &                        &   &\makecell[c]{50}        &  \makecell[c]{38M}   & 62.22    \\
             & ResNet-50         & \checkmark                       & \checkmark                      &  \checkmark                       &  \checkmark                       &    & \makecell[c]{50}    &  \makecell[c]{41M}    & 66.10    \\
             &  ResNet-50         &  \checkmark                       &   \checkmark                      & \checkmark                        &  \checkmark                      &\checkmark   & \makecell[c]{50}      &  \makecell[c]{42M}    & 68.65    \\
             &   ResNet-101        &               &           &              &                        &    &\makecell[c]{50}       & \makecell[c]{59M}   & 64.32    \\            
             &   ResNet-101        &   \checkmark                      & \checkmark                        & \checkmark                        &  \checkmark                      &    & \makecell[c]{50}        &  \makecell[c]{62M}     &  67.66   \\
             &  ResNet-101         &  \checkmark                       & \checkmark                        &  \checkmark                       &  \checkmark                      & \checkmark  &  \makecell[c]{50}       &   \makecell[c]{63M}    & 70.02    \\ \bottomrule
\end{tabular}
\end{center}
\caption{Comparisons with other baselines. (MS.downsample ratios mean downsampling ratio of multi-scale features)}
\label{tab1}
\end{table}

\subsection{Ablations}
\paragraph{Depthwise Separable Convolution.}
We evaluate the influence of Depthwise Separable Convolution (DSConv for abbreviation) by comparing encoder of DSConv and encoder of self-attention mechanism (Table~\ref{tab2}, $\rm O^2DETR$-Attn represents using self-attention mechanism in encoder and $\rm O^2DETR$-DSConv represents depthwise separable convolution). The implementation of $\rm O^2DETR$-Attn and $\rm O^2DETR$-DSConv keep the same except the components of encoder.
The number of layers for both DSConv and attention encoder keeps the same as 6 and all models adopt multi-scale features with the downsampling ratio of $64,32,16,8$. The results could prove the hypothesis that despite attention mechanisms use global scene reasoning in the whole image, in dense, tiny objects, local scene reasoning around objects is enough and even better, where local aggregation is the advantage of depthwise separable convolution. 

\begin{table}[h]
\centering
\begin{tabular}{l|llll}
\toprule
Method & backbone & Epochs & param & mAP \\ \midrule
$\rm O^2DETR$-Attn       & ResNet-50         &    \makecell[c]{50}   &  \makecell[c]{43M}  &  65.33   \\
$\rm O^2DETR$-Attn       & ResNet-101         &   \makecell[c]{50}   &  \makecell[c]{64M}  &  66.45   \\
$\rm O^2DETR$-DSConv       &   ResNet-50       &  \makecell[c]{50}   & \makecell[c]{41M}  &   66.10  \\ 
$\rm O^2DETR$-DSConv       &   ResNet-101      &  \makecell[c]{50}   &  \makecell[c]{62M} &   67.66 \\  \bottomrule
\end{tabular}
\caption{Comparisons between self-attention mechanism (\textbf{Transformer}) and depthwise separable convolution in encoder of $\rm O^2DETR$.}
\label{tab2}
\end{table}

\paragraph{Fine-tune on $\bf O^2DETR$.}
To be comparable with other detectors based on baseline of Faster R-CNN~\citep{ren2016faster} and RetinaNet~\citep{lin2017focal}, we propose to fine-tune the $\rm O^2DETR$ by utilizing $\rm O^2DETR$ as a region proposal network. The theoretical foundation of effectiveness of such fine-tune method is the high recall rate we report on Table~\ref{tab3}. We compute the recall of proposals at different IoU ratios with ground-truth boxes. The Recall-to-IoU metric is not strictly related to the ultimate detection accuracy, but it is an important metric to evaluate the proposal performance.
Compared with the RPN of Faster R-CNN~\citep{ren2016faster}, our model has higher recall rate, indicating our model covers more positive proposals. Different from original RPN, we keep the parameters of $\rm O^2DETR$ fixed to act as an inference network. The proposals predicted  by the inference network would be used into feature maps and finely modify bounding boxes and classifications to acquire better performance.

\begin{table}[h]
\centering
\begin{tabular}{l|llll}
\toprule
 & \multicolumn{4}{l}{\makecell[c]{IoU}} \\ \midrule
Method & \multicolumn{1}{l|}{\makecell[c]{0.2}} & \multicolumn{1}{l|}{\makecell[c]{0.3}} & \multicolumn{1}{l|}{\makecell[c]{0.4}} & \multicolumn{1}{l}{\makecell[c]{0.5}}  \\ \midrule
RPN   &\makecell[c]{47.86}   & \makecell[c]{44.22} & \makecell[c]{39.98} & \makecell[c]{35.11} \\
$\rm O^2DETR$    & \makecell[c]{68.49} &\makecell[c]{68.09}  & \makecell[c]{67.17} & \makecell[c]{65.27} \\ \bottomrule
\end{tabular}
\caption{Recall rates of RPN in Faster R-CNN~\citep{ren2016faster} and $\rm O^2DETR$.}
\label{tab3}
\end{table}

\begin{table}[h]
\centering
\begin{tabular}{l|llll}
\toprule
Method & backbone & Epochs & Fine-tune Epochs  & mAP \\ \midrule
 $\rm O^2DETR$       & ResNet-50         &   \makecell[c]{50}  &  \makecell[c]{0}     &  66.10   \\
$\rm O^2DETR$       & ResNet-101         &    \makecell[c]{50}   &  \makecell[c]{0}    &  67.66   \\
F-$\rm O^2DETR$       &   ResNet-50       &   \makecell[c]{50}    &   \makecell[c]{12}   &  74.47   \\ 
F-$\rm O^2DETR$       &   ResNet-101      &   \makecell[c]{50}    &   \makecell[c]{12}    &  76.23  \\  \bottomrule
\end{tabular}
\caption{Effectiveness of fine-tune.}
\label{tab4}
\end{table}

As is shown in Table~\ref{tab4}, the fine-tuning method improves the mAP performance largely by adding negligible extra fine-tuning epochs. All the models shown in Table~\ref{tab4} adopt multi-scale features with downsampling ratio $64,32,16,8$. The boost of performance demonstrates using $\rm O^2DETR$ as the baseline is feasible and has great potential. Other attempts are welcomed to be introduced to raise detection accuracy based on our baseline.

\subsection{Comparison with the State-of-the-art}
In this section, we compare our proposed $\rm O^2DETR$ with other state-of-the-art methods on an aerial oriented object detection dataset DOTA. The settings of our model have been introduced in Implementation Details. We achieve $74.47\%$ and $79.66\%$ mAP with ResNet-50-FPN backbone by fine-tuning the $\rm O^2DETR$ using single-scale and multi-scale dataset, respectively. In Table~\ref{tab5}, we report specific mAP performance in each categories (PL-plane, BD-baseball diamond, BR-bridge, GTF-ground track field, SV-small vehicle, LV-large vehicle, SH-ship, TC-tennis court, BC-basketball court, ST-storage tank, SBF-soccer ball field, RA-roundabout, HA-harbor, SP-swimming pool, HC-helicopter).

\begin{table}[h]
\scriptsize
\setlength{\tabcolsep}{0.6mm}
\begin{tabular}{l|l|lllllllllllllll|l}
\toprule
Method         & backbone & \makecell[c]{PL} & \makecell[c]{BD} & \makecell[c]{BR} & \makecell[c]{GTF} & \makecell[c]{SV} & \makecell[c]{LV} & \makecell[c]{SH} & \makecell[c]{TC} & \makecell[c]{BC} & \makecell[c]{ST} & \makecell[c]{SBF} & \makecell[c]{RA} & \makecell[c]{HA} & \makecell[c]{SP} & \makecell[c]{HC} & mAP \\ \midrule
single-scale   &          &    &    &    &     &    &    &    &    &    &    &     &    &    &    &    &     \\
FR-O           &R101     & 79.42   & 77.13   & 17.70   & 64.05    & 35.50   & 38.02   & 37.16   & 89.41   & 69.64   &59.28    & 50.30    & 52.91   & 47.89   &  47.40  & 46.30   & 54.13    \\
ICN            &R101-FPN  & 81.36   &74.30    & 47.70   &70.32     &64.89    & 67.82   & 69.98   & 90.76   & 79.06   & 78.20   & 53.64    & 62.90   & 67.02   & 64.17   & 50.23   &  68.16   \\
CADNet         & R101-FPN         & 87.80   & 82.40   & 49.40   & 73.50    &71.10    & 63.50   & 76.60   &90.90    & 79.20   & 73.30   &48.40     &60.90    &62.00    &67.00    & 62.20   & 69.90    \\
DRN            &  H-104        & 88.91   & 80.22   &43.52    & 63.35    & 73.48   &70.69    &84.94    &90.14    &83.85    &84.11    & 50.12    &58.41    & 67.62   & 68.60   &52.50    & 70.70    \\
CenterMap      &  R50-FPN        & 88.88   & 81.24   & 53.15   & 60.65    & 78.62   &66.55    &78.10    & 88.83   &77.80    & 83.61   &49.36     & 66.19   & 72.10   &72.36    & 58.70   & 71.74    \\
SCRDet         &  R101-FPN        & 89.98   & 80.65   & 52.09   &68.36   &68.36  & 60.32   & 72.41   & 90.85   & 87.94   & 86.86   &65.02    &  66.68   & 66.25   &  68.24  & 65.21      & 72.61    \\
$\rm R^3Det$   &  R152-FPN        &89.49    & 81.17   & 50.53   & 66.10    & 70.92   &78.66    & 78.21   & 90.81   &  85.26  & 84.23   & 61.81    & 63.77   & 68.16   & 69.83   &67.17    & 73.74    \\
$\rm S^2ANet$      & R50-FPN         &89.11    & 82.84   & 48.37   &71.11     & 78.11   &78.39    & 87.25   &  90.83  &  84.90  &85.64    &  60.36   & 62.60   &65.26    & 69.13   &  57.94  & 74.12    \\
ReDet          &   ReR50-ReFPN       &  88.79  &  82.64  &53.97    & 74.00    & 78.13   & 84.06   & 88.04   & 90.89   & 87.78   & 85.75   & 61.76    &60.39   &75.96   &  68.07  &63.59    &  76.25   \\
$\rm O^2DETR$      & R50-FPN         & 83.89   &75.11    & 44.04   & 64.20    & 78.39   & 76.78   & 87.68   & 90.60   & 78.58   &71.82    &53.21     & 60.35   &55.36    &61.90    & 47.89   & 68.65    \\
F-$\rm O^2DETR$    &  R50-FPN        & 88.76   & 81.91   & 51.20   &72.18     &77.64    & 80.47   &87.84    & 90.85   &  84.56  & 81.68   & 61.42    &64.61    & 67.50   & 64.28   & 62.15   & 74.47    \\ \midrule
multi-scale    &          &    &    &    &     &    &    &    &    &    &    &     &    &    &    &    &     \\
ROI Trans      &  R101-FPN       &88.64 &78.52 &43.44 &75.92 &68.81 &73.68 &83.59 &90.74 &77.27 &81.46 &58.39 &53.54 &62.83 &58.93 &47.67 &69.56   \\
$\rm O^2$-DNet     &  H104       &89.30 &83.30 &50.10 &72.10 &71.10 &75.60 &78.70 &90.90 &79.90 &82.90 &60.20 &60.00 &64.60 &68.90 &65.70    & 72.80    \\
DRN            &  H104        &89.71 &82.34 &47.22 &64.10 &76.22 &74.43 &85.84 &90.57 &86.18 &84.89 &57.65 &61.93 &69.30 &69.63 &58.48   & 73.23    \\
Gliding Vertex & R101-FPN   &89.64 &85.00 &52.26 &77.34 &73.01 &73.14 &86.82 &90.74 &79.02 &86.81 &59.55 &70.91 &72.94 &70.86 &57.32 &75.02   \\
BBAVectors     &  R101  &88.63 &84.06 &52.13 &69.56 &78.26 &80.40 &88.06 &90.87 &87.23 &86.39 &56.11 &65.62 &67.10 &72.08 &63.96 &75.36   \\
CenterMap      & R101-FPN        &89.83 &84.41 &54.60 &70.25 &77.66 &78.32 &87.19 &90.66 &84.89 &85.27 &56.46 &69.23 &74.13 &71.56 &66.06    & 76.03    \\
CSL            &  R152-FPN       &90.25 &85.53 &54.64 &75.31 &70.44 &73.51 &77.62 &90.84 &86.15 &86.69 &69.60 &68.04 &73.83 &71.10 &68.93    & 76.17    \\
SCRDet++       &  R152-FPN        &88.68 &85.22 &54.70 &73.71 &71.92 &84.14 &79.39 &90.82 &87.04 &86.02 &67.90 &60.86 &74.52 &70.76 &72.66    &76.56     \\
$\rm S^2ANet$      &  R50-FPN      & 88.89 &83.60 &57.74 &81.95 &79.94 &83.19 &89.11 &90.78 &84.87 &87.81 &70.30 &68.25 &78.30 &77.01 &69.58   & 79.42    \\
ReDet          &   ReR50-ReFPN      &88.81 &82.48 &60.83 &80.82 &78.34 &86.06 &88.31 &90.87 &88.77 &87.03 &68.65 &66.90 &79.26 &79.71 &74.67    &80.10     \\
$\rm O^2DETR$      & R50-FPN         &86.01    &75.92    & 46.02   &66.65     & 79.70  & 79.93 & 89.17   & 90.44   &  81.19  & 76.00   & 56.91   & 62.45    & 64.22   & 65.80   & 58.96   &  72.15       \\
F-$\rm O^2DETR$    & R50-FPN         & 88.89   &  83.41  & 56.72   & 79.75    & 79.89   & 85.45   & 89.77   & 90.84   &  86.15  & 87.66   & 69.94    & 68.97   & 78.83   &  78.19  & 70.38   & 79.66    \\ \bottomrule
\end{tabular}
\caption{Comparison with state-of-the-art methods on DOTA. R-50(101)-FPN stands for ResNet-50(101) with FPN and H104 stands for Hourglass-104. F-$\rm O^2DETR$ means fine-tuned $\rm O^2DETR$. Multi-scale indicates training and testing on multi-scale cropped images.}
\label{tab5}
\end{table}

By simply fine-tuning the $\rm O^2DETR$, the performance is quite competitive compared with SOTA performance. It illustrates our fine-tuned model performs better than $\rm S^2ANet$~\citep{han2020align} as backbone is ResNet-50 with FPN. The SOTA performance, ReDet~\citep{han2021redet}, retraining models backbone based on ResNet-50 and is not strictly comparable with other our model and $\rm S^2ANet$. 

\begin{figure}
    \centering
    \includegraphics[width=1\linewidth]{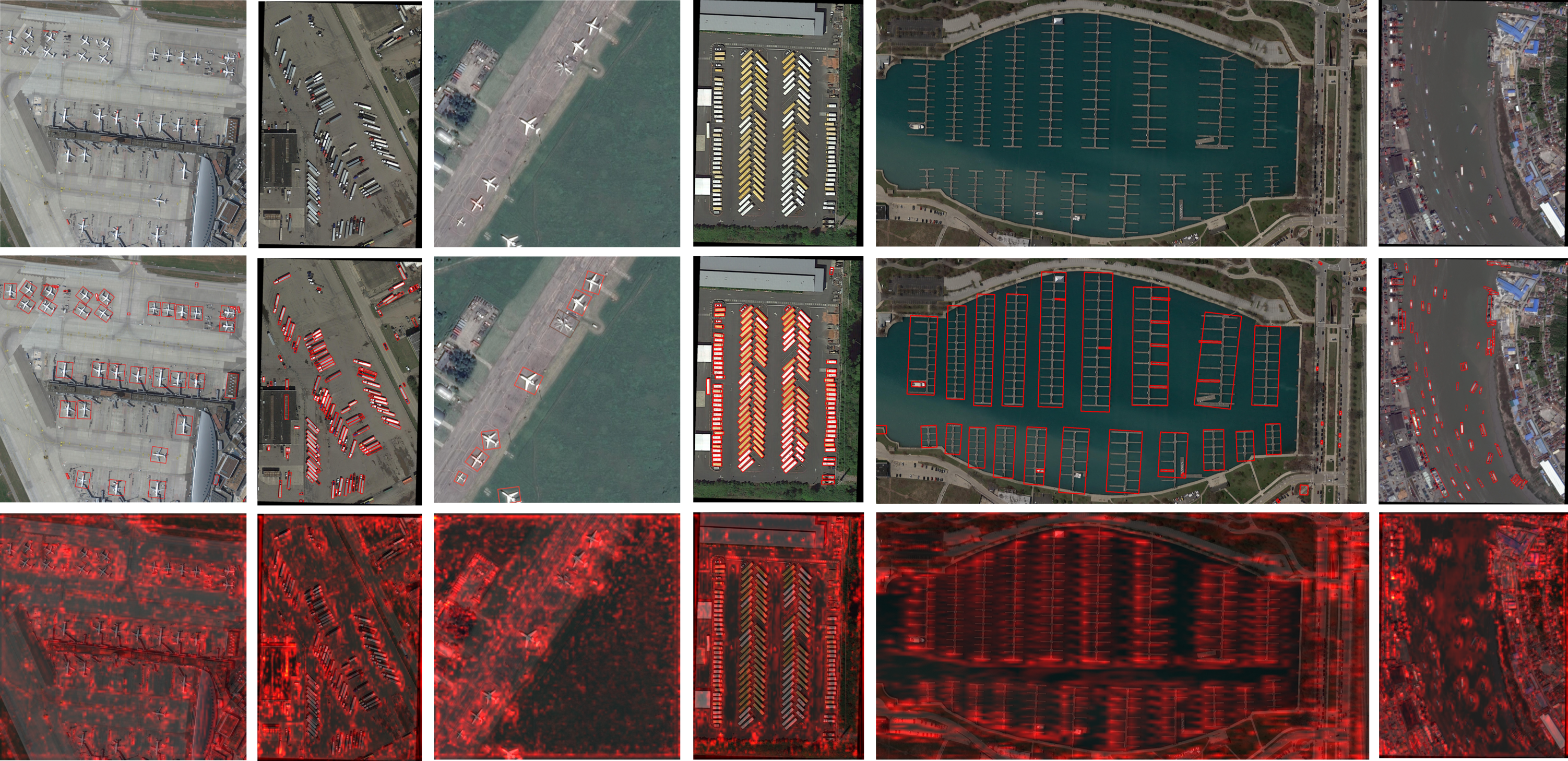}
    \caption{Visualization of detecting with $\rm O^2DETR$. The first row is images to be detected, and the second row is detection results of $\rm O^2DETR$. The third row is visualization of depth-wise encoding.}
    \label{fig3}
\end{figure}

We show some qualitative samples in Fig.~\ref{fig3}. The visualization of detpth-wise encoding in the third row of Fig.~\ref{fig3}. We visualize the activation map of encoded features after depthwise convolution. The activation map will emphasize dense object region in input image. Compared with the global reasoning of attention mechanism, the local aggregation is enough to refer the targets in a less complex way.

\section{Conclusion}
In this paper, we propose a new end-to-end  model, $\rm O^2DETR$, for oriented object detection problem via Transformer. The $\rm O^2DETR$ outperforms original rotated Faster R-CNN and RetinaNet baseline on the challenging DOTA dataset. The $\rm O^2DETR$ is straightforward and flexible to apply in oriented object detection. Based on that, we provide a new method by fine-tuning $\rm O^2DETR$ to achieve a competitive performance compared with the state-of-the-arts. Extensive ablations on DOTA dataset demonstrate the effectiveness of our method. In the future work, we will try more applications to verify the performance of our method.

{
\bibliographystyle{plainnat}
\bibliography{infer}

\begin{thebibliography}{41}
\providecommand{\natexlab}[1]{#1}
\providecommand{\url}[1]{\texttt{#1}}
\expandafter\ifx\csname urlstyle\endcsname\relax
  \providecommand{\doi}[1]{doi: #1}\else
  \providecommand{\doi}{doi: \begingroup \urlstyle{rm}\Url}\fi

\bibitem[Beltagy et~al.(2020)Beltagy, Peters, and Cohan]{beltagy2020longformer}
Iz~Beltagy, Matthew~E Peters, and Arman Cohan.
\newblock Longformer: The long-document transformer.
\newblock \emph{arXiv preprint arXiv:2004.05150}, 2020.

\bibitem[Brown et~al.(2020)Brown, Mann, Ryder, Subbiah, Kaplan, Dhariwal,
  Neelakantan, Shyam, Sastry, Askell, et~al.]{brown2020language}
Tom~B Brown, Benjamin Mann, Nick Ryder, Melanie Subbiah, Jared Kaplan, Prafulla
  Dhariwal, Arvind Neelakantan, Pranav Shyam, Girish Sastry, Amanda Askell,
  et~al.
\newblock Language models are few-shot learners.
\newblock \emph{arXiv preprint arXiv:2005.14165}, 2020.

\bibitem[Carion et~al.(2020)Carion, Massa, Synnaeve, Usunier, Kirillov, and
  Zagoruyko]{carion2020end}
Nicolas Carion, Francisco Massa, Gabriel Synnaeve, Nicolas Usunier, Alexander
  Kirillov, and Sergey Zagoruyko.
\newblock End-to-end object detection with transformers.
\newblock In \emph{European Conference on Computer Vision}, pages 213--229,
  2020.

\bibitem[Chollet(2017)]{chollet2017xception}
Fran{\c{c}}ois Chollet.
\newblock Xception: Deep learning with depthwise separable convolutions.
\newblock In \emph{Proceedings of the IEEE/CVF Conference on Computer Vision
  and Pattern Recognition}, pages 1251--1258, 2017.

\bibitem[Deng et~al.(2009)Deng, Dong, Socher, Li, Li, and
  Fei-Fei]{deng2009imagenet}
Jia Deng, Wei Dong, Richard Socher, Li-Jia Li, Kai Li, and Li~Fei-Fei.
\newblock Imagenet: A large-scale hierarchical image database.
\newblock In \emph{Proceedings of the IEEE/CVF Conference on Computer Vision
  and Pattern Recognition}, 2009.

\bibitem[Devlin et~al.(2018)Devlin, Chang, Lee, and Toutanova]{devlin2018bert}
Jacob Devlin, Ming-Wei Chang, Kenton Lee, and Kristina Toutanova.
\newblock Bert: Pre-training of deep bidirectional transformers for language
  understanding.
\newblock \emph{arXiv preprint arXiv:1810.04805}, 2018.

\bibitem[Ding et~al.(2019)Ding, Xue, Long, Xia, and Lu]{ding2019learning}
Jian Ding, Nan Xue, Yang Long, Gui-Song Xia, and Qikai Lu.
\newblock Learning roi transformer for oriented object detection in aerial
  images.
\newblock In \emph{Proceedings of the IEEE/CVF Conference on Computer Vision
  and Pattern Recognition}, pages 2849--2858, 2019.

\bibitem[Dosovitskiy et~al.(2020)Dosovitskiy, Beyer, Kolesnikov, Weissenborn,
  Zhai, Unterthiner, Dehghani, Minderer, Heigold, Gelly,
  et~al.]{dosovitskiy2020image}
Alexey Dosovitskiy, Lucas Beyer, Alexander Kolesnikov, Dirk Weissenborn,
  Xiaohua Zhai, Thomas Unterthiner, Mostafa Dehghani, Matthias Minderer, Georg
  Heigold, Sylvain Gelly, et~al.
\newblock An image is worth 16$\times$ 16 words: Transformers for image
  recognition at scale.
\newblock 2020.

\bibitem[Gao et~al.(2019{\natexlab{a}})Gao, Jiang, You, Lu, Hoi, Wang, and
  Li]{gao2019dynamic}
Peng Gao, Zhengkai Jiang, Haoxuan You, Pan Lu, Steven~CH Hoi, Xiaogang Wang,
  and Hongsheng Li.
\newblock Dynamic fusion with intra-and inter-modality attention flow for
  visual question answering.
\newblock In \emph{Proceedings of the IEEE/CVF Conference on Computer Vision
  and Pattern Recognition}, pages 6639--6648, 2019{\natexlab{a}}.

\bibitem[Gao et~al.(2019{\natexlab{b}})Gao, You, Zhang, Wang, and
  Li]{gao2019multi}
Peng Gao, Haoxuan You, Zhanpeng Zhang, Xiaogang Wang, and Hongsheng Li.
\newblock Multi-modality latent interaction network for visual question
  answering.
\newblock In \emph{Proceedings of the IEEE/CVF International Conference on
  Computer Vision}, pages 5825--5835, 2019{\natexlab{b}}.

\bibitem[Gao et~al.(2020)Gao, Hori, Geng, Hori, and Roux]{gao2020multi}
Peng Gao, Chiori Hori, Shijie Geng, Takaaki Hori, and Jonathan~Le Roux.
\newblock Multi-pass transformer for machine translation.
\newblock \emph{arXiv preprint arXiv:2009.11382}, 2020.

\bibitem[Gao et~al.(2021)Gao, Zheng, Wang, Dai, and Li]{gao2021fast}
Peng Gao, Minghang Zheng, Xiaogang Wang, Jifeng Dai, and Hongsheng Li.
\newblock Fast convergence of detr with spatially modulated co-attention.
\newblock \emph{arXiv preprint arXiv:2101.07448}, 2021.

\bibitem[Geng et~al.(2020)Geng, Gao, Chatterjee, Hori, Roux, Zhang, Li, and
  Cherian]{geng2020dynamic}
Shijie Geng, Peng Gao, Moitreya Chatterjee, Chiori Hori, Jonathan~Le Roux,
  Yongfeng Zhang, Hongsheng Li, and Anoop Cherian.
\newblock Dynamic graph representation learning for video dialog via
  multi-modal shuffled transformers.
\newblock \emph{arXiv preprint arXiv:2007.03848}, 2020.

\bibitem[Han et~al.(2020)Han, Ding, Li, and Xia]{han2020align}
Jiaming Han, Jian Ding, Jie Li, and Gui-Song Xia.
\newblock Align deep features for oriented object detection.
\newblock \emph{arXiv preprint arXiv:2008.09397}, 2020.

\bibitem[Han et~al.(2021)Han, Ding, Xue, and Xia]{han2021redet}
Jiaming Han, Jian Ding, Nan Xue, and Gui-Song Xia.
\newblock Redet: A rotation-equivariant detector for aerial object detection.
\newblock \emph{arXiv preprint arXiv:2103.07733}, 2021.

\bibitem[He et~al.(2016)He, Zhang, Ren, and Sun]{he2016deep}
Kaiming He, Xiangyu Zhang, Shaoqing Ren, and Jian Sun.
\newblock Deep residual learning for image recognition.
\newblock In \emph{Proceedings of the IEEE/CVF Conference on Computer Vision
  and Pattern Recognition}, pages 770--778, 2016.

\bibitem[Howard et~al.(2017)Howard, Zhu, Chen, Kalenichenko, Wang, Weyand,
  Andreetto, and Adam]{howard2017mobilenets}
Andrew~G Howard, Menglong Zhu, Bo~Chen, Dmitry Kalenichenko, Weijun Wang,
  Tobias Weyand, Marco Andreetto, and Hartwig Adam.
\newblock Mobilenets: Efficient convolutional neural networks for mobile vision
  applications.
\newblock \emph{arXiv preprint arXiv:1704.04861}, 2017.

\bibitem[Jiang et~al.(2017)Jiang, Zhu, Wang, Yang, Li, Wang, Fu, and
  Luo]{jiang2017r2cnn}
Yingying Jiang, Xiangyu Zhu, Xiaobing Wang, Shuli Yang, Wei Li, Hua Wang, Pei
  Fu, and Zhenbo Luo.
\newblock R2cnn: rotational region cnn for orientation robust scene text
  detection.
\newblock \emph{arXiv preprint arXiv:1706.09579}, 2017.

\bibitem[Katharopoulos et~al.(2020)Katharopoulos, Vyas, Pappas, and
  Fleuret]{katharopoulos2020transformers}
Angelos Katharopoulos, Apoorv Vyas, Nikolaos Pappas, and Fran{\c{c}}ois
  Fleuret.
\newblock Transformers are rnns: Fast autoregressive transformers with linear
  attention.
\newblock In \emph{International Conference on Machine Learning}, pages
  5156--5165. PMLR, 2020.

\bibitem[Kitaev et~al.(2020)Kitaev, Kaiser, and Levskaya]{kitaev2020reformer}
Nikita Kitaev, {\L}ukasz Kaiser, and Anselm Levskaya.
\newblock Reformer: The efficient transformer.
\newblock \emph{arXiv preprint arXiv:2001.04451}, 2020.

\bibitem[Lin et~al.(2017{\natexlab{a}})Lin, Doll{\'a}r, Girshick, He,
  Hariharan, and Belongie]{lin2017feature}
Tsung-Yi Lin, Piotr Doll{\'a}r, Ross Girshick, Kaiming He, Bharath Hariharan,
  and Serge Belongie.
\newblock Feature pyramid networks for object detection.
\newblock In \emph{Proceedings of the IEEE/CVF Conference on Computer Vision
  and Pattern Recognition}, pages 2117--2125, 2017{\natexlab{a}}.

\bibitem[Lin et~al.(2017{\natexlab{b}})Lin, Goyal, Girshick, He, and
  Doll{\'a}r]{lin2017focal}
Tsung-Yi Lin, Priya Goyal, Ross Girshick, Kaiming He, and Piotr Doll{\'a}r.
\newblock Focal loss for dense object detection.
\newblock In \emph{Proceedings of the IEEE international conference on computer
  vision}, pages 2980--2988, 2017{\natexlab{b}}.

\bibitem[Lu et~al.(2019)Lu, Batra, Parikh, and Lee]{lu2019vilbert}
Jiasen Lu, Dhruv Batra, Devi Parikh, and Stefan Lee.
\newblock Vilbert: Pretraining task-agnostic visiolinguistic representations
  for vision-and-language tasks.
\newblock 2019.

\bibitem[Ma et~al.(2018)Ma, Shao, Ye, Wang, Wang, Zheng, and
  Xue]{ma2018arbitrary}
Jianqi Ma, Weiyuan Shao, Hao Ye, Li~Wang, Hong Wang, Yingbin Zheng, and
  Xiangyang Xue.
\newblock Arbitrary-oriented scene text detection via rotation proposals.
\newblock \emph{IEEE Transactions on Multimedia}, 20\penalty0 (11):\penalty0
  3111--3122, 2018.

\bibitem[Mao et~al.(2021)Mao, Zhang, Zheng, Gao, Ma, Peng, Ding, and
  Han]{mao2021dual}
Mingyuan Mao, Renrui Zhang, Honghui Zheng, Peng Gao, Teli Ma, Yan Peng, Errui
  Ding, and Shumin Han.
\newblock Dual-stream network for visual recognition.
\newblock \emph{arXiv preprint arXiv:2105.14734}, 2021.

\bibitem[Ott et~al.(2018)Ott, Edunov, Grangier, and Auli]{ott2018scaling}
Myle Ott, Sergey Edunov, David Grangier, and Michael Auli.
\newblock Scaling neural machine translation.
\newblock \emph{arXiv preprint arXiv:1806.00187}, 2018.

\bibitem[Radford et~al.(2018)Radford, Narasimhan, Salimans, and
  Sutskever]{radford2018improving}
Alec Radford, Karthik Narasimhan, Tim Salimans, and Ilya Sutskever.
\newblock Improving language understanding by generative pre-training.
\newblock 2018.

\bibitem[Radford et~al.(2019)Radford, Wu, Child, Luan, Amodei, and
  Sutskever]{radford2019language}
Alec Radford, Jeffrey Wu, Rewon Child, David Luan, Dario Amodei, and Ilya
  Sutskever.
\newblock Language models are unsupervised multitask learners.
\newblock \emph{OpenAI blog}, 1\penalty0 (8):\penalty0 9, 2019.

\bibitem[Ramachandran et~al.(2019)Ramachandran, Parmar, Vaswani, Bello,
  Levskaya, and Shlens]{ramachandran2019stand}
Prajit Ramachandran, Niki Parmar, Ashish Vaswani, Irwan Bello, Anselm Levskaya,
  and Jonathon Shlens.
\newblock Stand-alone self-attention in vision models.
\newblock \emph{arXiv preprint arXiv:1906.05909}, 2019.

\bibitem[Ren et~al.(2016)Ren, He, Girshick, and Sun]{ren2016faster}
Shaoqing Ren, Kaiming He, Ross Girshick, and Jian Sun.
\newblock Faster r-cnn: towards real-time object detection with region proposal
  networks.
\newblock \emph{IEEE transactions on pattern analysis and machine
  intelligence}, 39\penalty0 (6):\penalty0 1137--1149, 2016.

\bibitem[Sifre and Mallat(2013)]{sifre2013rotation}
Laurent Sifre and St{\'e}phane Mallat.
\newblock Rotation, scaling and deformation invariant scattering for texture
  discrimination.
\newblock In \emph{Proceedings of the IEEE/CVF Conference on Computer Vision
  and Pattern Recognition}, pages 1233--1240, 2013.

\bibitem[Vaswani et~al.(2017)Vaswani, Shazeer, Parmar, Uszkoreit, Jones, Gomez,
  Kaiser, and Polosukhin]{vaswani2017attention}
Ashish Vaswani, Noam Shazeer, Niki Parmar, Jakob Uszkoreit, Llion Jones,
  Aidan~N Gomez, Lukasz Kaiser, and Illia Polosukhin.
\newblock Attention is all you need.
\newblock In \emph{NIPS}, 2017.

\bibitem[Wang et~al.(2020)Wang, Li, Khabsa, Fang, and Ma]{wang2020linformer}
Sinong Wang, Belinda Li, Madian Khabsa, Han Fang, and Hao Ma.
\newblock Linformer: Self-attention with linear complexity.
\newblock \emph{arXiv preprint arXiv:2006.04768}, 2020.

\bibitem[Xia et~al.(2018)Xia, Bai, Ding, Zhu, Belongie, Luo, Datcu, Pelillo,
  and Zhang]{xia2018dota}
Gui-Song Xia, Xiang Bai, Jian Ding, Zhen Zhu, Serge Belongie, Jiebo Luo, Mihai
  Datcu, Marcello Pelillo, and Liangpei Zhang.
\newblock Dota: A large-scale dataset for object detection in aerial images.
\newblock In \emph{Proceedings of the IEEE/CVF Conference on Computer Vision
  and Pattern Recognition}, pages 3974--3983, 2018.

\bibitem[Yang and Yan(2020)]{yang2020arbitrary}
Xue Yang and Junchi Yan.
\newblock Arbitrary-oriented object detection with circular smooth label.
\newblock In \emph{European Conference on Computer Vision}, pages 677--694,
  2020.

\bibitem[Yang et~al.(2019{\natexlab{a}})Yang, Liu, Yan, Li, Zhang, and
  Yu]{yang2019r3det}
Xue Yang, Qingqing Liu, Junchi Yan, Ang Li, Zhiqiang Zhang, and Gang Yu.
\newblock R3det: Refined single-stage detector with feature refinement for
  rotating object.
\newblock \emph{arXiv preprint arXiv:1908.05612}, 2019{\natexlab{a}}.

\bibitem[Yang et~al.(2019{\natexlab{b}})Yang, Yang, Yan, Zhang, Zhang, Guo,
  Sun, and Fu]{yang2019scrdet}
Xue Yang, Jirui Yang, Junchi Yan, Yue Zhang, Tengfei Zhang, Zhi Guo, Xian Sun,
  and Kun Fu.
\newblock Scrdet: Towards more robust detection for small, cluttered and
  rotated objects.
\newblock In \emph{Proceedings of the IEEE International Conference on Computer
  Vision}, pages 8232--8241, 2019{\natexlab{b}}.

\bibitem[Yu et~al.(2019)Yu, Yu, Cui, Tao, and Tian]{yu2019deep}
Zhou Yu, Jun Yu, Yuhao Cui, Dacheng Tao, and Qi~Tian.
\newblock Deep modular co-attention networks for visual question answering.
\newblock In \emph{Proceedings of the IEEE/CVF Conference on Computer Vision
  and Pattern Recognition}, pages 6281--6290, 2019.

\bibitem[Zheng et~al.(2020)Zheng, Gao, Wang, Li, and Dong]{zheng2020end}
Minghang Zheng, Peng Gao, Xiaogang Wang, Hongsheng Li, and Hao Dong.
\newblock End-to-end object detection with adaptive clustering transformer.
\newblock \emph{arXiv preprint arXiv:2011.09315}, 2020.

\bibitem[Zhou et~al.(2020)Zhou, Wei, Li, Zhao, Zhang, and
  Zhang]{zhou2020objects}
Lin Zhou, Haoran Wei, Hao Li, Wenzhe Zhao, Yi~Zhang, and Yue Zhang.
\newblock Objects detection for remote sensing images based on polar
  coordinates.
\newblock \emph{arXiv preprint arXiv:2001.02988}, 2020.

\bibitem[Zhu et~al.(2020)Zhu, Su, Lu, Li, Wang, and Dai]{zhu2020deformable}
Xizhou Zhu, Weijie Su, Lewei Lu, Bin Li, Xiaogang Wang, and Jifeng Dai.
\newblock Deformable detr: Deformable transformers for end-to-end object
  detection.
\newblock \emph{arXiv preprint arXiv:2010.04159}, 2020.

\end{thebibliography}
}



\end{document}